\pdfoutput=1

\documentclass[11pt]{article}

\usepackage[final]{acl}

\usepackage{times}
\usepackage{latexsym}

\usepackage[T1]{fontenc}

\usepackage[utf8]{inputenc}
\usepackage{enumitem}
\usepackage{microtype}
\usepackage{amssymb}
\usepackage{amsmath}
\usepackage{bbm}
\usepackage{bm}

\usepackage{inconsolata}

\usepackage{graphicx}
\usepackage{booktabs}

\usepackage{tabularx}
\usepackage{multirow}

\usepackage{float} 
\usepackage{colortbl}
\usepackage{setspace}
\usepackage[accsupp]{axessibility}

\title{MaPPER: Multimodal Prior-guided Parameter Efficient Tuning for Referring Expression Comprehension}

\author{
 \textbf{Ting Liu\textsuperscript{1}},
 \textbf{Zunnan Xu\textsuperscript{2}},
 \textbf{Yue Hu\textsuperscript{1}},
 \textbf{Liangtao Shi\textsuperscript{3}},
 \textbf{Zhiqiang Wang\textsuperscript{4}},
 \textbf{Quanjun Yin\textsuperscript{1}\footnotemark[2]}
\\
 \textsuperscript{1}College of Systems Engineering, National University of Defense Technology,\\
 \textsuperscript{2}Tsinghua University,
 \textsuperscript{3}Hefei University of Technology,
 \textsuperscript{4}iFLYTEK Research\\
\tt\small liuting20@nudt.edu.cn
}

\begin{document}
\maketitle

\def\thefootnote{\dag}\footnotetext{Corresponding author}
\def\thefootnote{\arabic{footnote}}

\begin{abstract}
\label{sec:abstract}
Referring Expression Comprehension (REC), which aims to ground a local visual region via natural language, is a task that heavily relies on multimodal alignment. Most existing methods utilize powerful pre-trained models to transfer visual/linguistic knowledge by full fine-tuning. However, full fine-tuning the entire backbone not only breaks the rich prior knowledge embedded in the pre-training, but also incurs significant computational costs. Motivated by the recent emergence of Parameter-Efficient Transfer Learning (PETL) methods, we aim to solve the REC task in an effective and efficient manner. Directly applying these PETL methods to the REC task is inappropriate, as they lack the specific-domain abilities for precise local visual perception and visual-language alignment. Therefore, we propose a novel framework of Multimodal Prior-guided Parameter Efficient Tuning, namely MaPPER. Specifically, MaPPER comprises Dynamic Prior Adapters guided by an aligned prior, and Local Convolution Adapters to extract precise local semantics for better visual perception. Moreover, the Prior-Guided Text module is proposed to further utilize the prior for facilitating the cross-modal alignment. Experimental results on three widely-used benchmarks demonstrate that MaPPER achieves the best accuracy compared to the full fine-tuning and other PETL methods with only \textbf{1.41\%} tunable backbone parameters. Our code is available at \url{https://github.com/liuting20/MaPPER}.

\end{abstract}
\section{Introduction}
\label{sec:intro}

\begin{figure}[t]
\centering
\includegraphics[width=1.0\columnwidth]{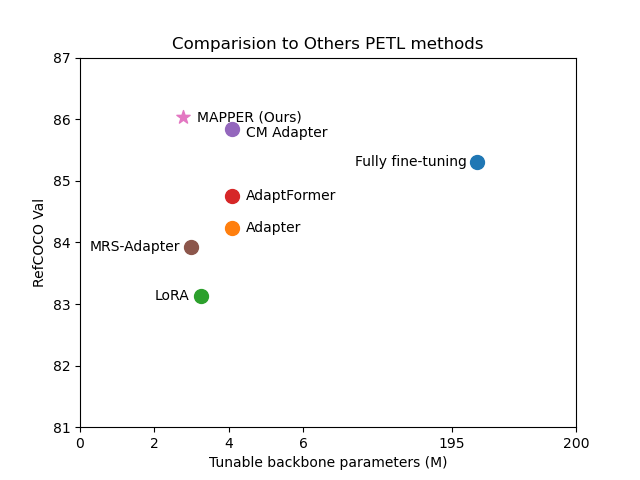}
\caption{Comparision to others PETL methods.
}
\label{fig:other-petl data}
\end{figure}

Referring Expression Comprehension (REC)~\cite{kamath2021mdetr,liu2023dqdetr,wu2023scene,bu2023KB-REC} is a crucial and challenging task within the multimodal fields, which needs to localize the local image region according to the language expression semantics. REC is fundamental for visual language understanding, with broad applications in fields such as visual-language navigation~\cite{liu2024dap} and human-machine interaction~\cite{chen2023shikra}.
Different from vanilla object detection task, REC needs to extract not only global and local spatial information from images, but also relies on the alignment of multimodal features.

Existing approaches~\cite{deng2021transvg,kamath2021mdetr,transvg++,shi2023dynamic} transfer the language and vision knowledge from pre-trained models by fully fine-tuning. However, such a fine-tuning strategy is sub-optimal for REC, as reflected in the following aspects: \textbf{1)} Fine-tuning the entire backbone might suffer catastrophic forgetting and undermine the extensive prior knowledge learned from pre-training. \textbf{2)} The computational cost requirements surge dramatically, particularly for larger foundational models, leading to a significant increase in GPU memory usage. This limits the accessibility of large models for researchers with limited hardware resources.

To address these issues, we shift our focus to Parameter-Efficient Transfer Learning (PETL)~\cite{chowdhury2023apollo,wang2023aprompt}. PETL methods like Adapter tuning and Prompt tuning provide efficient ways to utilize pre-trained models by adjusting a small set of parameters instead of fine-tuning the entire network~\cite{xin2024parameter}. This approach saves computational resources while still competitive performance improvements. By integrating PETL techniques, we can enhance our models' flexibility and efficiency in adapting to REC. However, we empirically find that directly using these PETL methods cannot achieve satisfactory results in REC (see Figure~\ref{fig:other-petl data}).
We argue the main reasons are twofold: \textbf{1)} the target objects that require attention in REC often occupy local regions of uncertain size in images, and most existing PETL methods lack the crucial ability to extract multi-scale local semantics for visual perception. \textbf{2)} REC is a task that strongly relies on  multimodal alignment, and language-oriented adapters are obviously deficient in aligning with visual information. Recently, PETL methods have also been introduced into REC tasks~\cite{xiao2024hivg,liu2024dara}.
HiVG~\cite{xiao2024hivg} adopts LoRA to fine-tune the frozen CLIP model, but it is not an efficient enough approach due to the heavy alignment design using cross-attention module. In contrast, DARA~\cite{liu2024dara} is a lightweight method in PETL paradigm. However, DARA does not fully address the need for local visual adaptation in the referring expression comprehension, potentially compromising the model's ability to capture fine-grained visual details.

Considering the aforementioned issues, in this paper, we propose a novel framework of \textbf{M}ultimod\textbf{a}l \textbf{P}rior-guided \textbf{P}arameter \textbf{E}fficient Tuning for \textbf{R}EC (MaPPER) that improves text understanding with the aligned prior and enhances vision perception by combining local visual semantics with global perception.
As shown in Figure~\ref{fig:overview}, we introduce the vision-aligned text module to generate the aligned prior, which works for the alignment of vision and language feature. Moreover, we insert the Local Convolution Adapter (LoCA) into vision blocks for enhancing visual perception. Specifically, we propose the Dynamic Prior Adapter (DyPA) presented in Figure~\ref{fig:dypa}, DyPA can dynamically adjust each token by considering the significance score guided by the aligned prior. In order to promote the interaction of text and vision features, we further propose the Prior-guided Text module (PGT) for fusing the prior and text feature. For the visual branch, most pre-trained visual models are powerful transformer architectures. Unfortunately, vision transformers are observed ignoring local feature details~\cite{peng2021conformer}, which decreases the discriminability between backgrounds and foregrounds. Motivated by this, we introduce the Local Convolution Adapter (LoCA), which integrates multi-scale local knowledge, thereby enhancing the representational power for pre-trained vision transformers.
Extensive experiments on RefCOCO~\cite{yu2016refcoco}, RefCOCO+~\cite{yu2016refcoco}, and RefCOCOg~\cite{mao2016refcocogg,nagaraja2016refcocogu} demonstrate the effectiveness and efficiency of our framework. 
Our main contributions are as follows:
\begin{itemize}[leftmargin=*,noitemsep,nolistsep]
    \item We perform an in-depth exploration of parameter-efficient transfer learning (PETL) methods for REC tasks. We introduce MaPPER aimed at improving both the effectiveness and efficiency of visual-text alignment, as well as enhancing visual perception by incorporating local visual semantics.

    \item We propose the novel Dynamic Prior Adapter (DyPA) and Local Convolution Adapter (LoCA). The former employs aligned prior to dynamically adjust the language encoder, while the latter introduces local visual features for enhancing the visual encoder.
    
    \item Extensive experiments demonstrate that our method can outperform the state-of-the-art (SOTA) methods in REC tasks, with only \textbf{1.41\%} tunable parameters within pre-trained backbones.
\end{itemize}

\section{Related Work}
\label{sec:related work}

\subsection{Referring Expression Comprehension}
Referring expression comprehension (REC)~\cite{yu2018mattnet,yang2019fast,deng2021transvg,xiao2023clip,liu2024vgdiffzero,xiao2024hivg} aims to locate a local visual region in images by textual descriptions. Early propose-and-rank methods~\cite{hong2019learning,chen2020referring} follow a two-stage pipeline which first utilizes pre-trained object detectors to obtain a set of region proposals, which are then ranked based on their similarity scores with the given textual description. However, these two-stage methods face challenges in terms of the performance of the proposal generators and the additional ranking mechanisms. After the introduction of ViT, the Transformer-based methods~\cite{deng2021transvg,du2022vgtr,yang2022vltvg,zhu2022seqtr,liu2024dara,zhu2023jmri} have recently emerged that significantly improve the grounding performance. Most recently, grounding multimodal large language models~\cite{li2023g2l,wang2023cogvlm} have propelled the state-of-the-art (SOTA) performance, these works require a large amount of in-domain and other domain datasets. As REC models continue to scale up in size and complexity, fully fine-tuning becomes extremely high training cost.

\subsection{Parameter-efficient Transfer Learning}

The continuous expansion of pre-trained models demands significant computational resources and consumes considerable storage during fine-tuning~\cite{liu2024sparse}. To address these challenges, researchers in the NLP and CV domain have explored PETL methods ~\cite{hu2021lora,chen2022adaptformer,yuan2023mrsadapter,liu2024panda}. By focusing on updating only a small subset of parameters, PETL achieves a balance between maintaining high performance and ensuring computational efficiency. This method is particularly advantageous for deploying large-scale models, addressing the challenges posed by increasing model sizes while streamlining the adaptation process to new tasks. The main PETL methods can be classified into three categories:
(i) selectively updating a tiny number of existing model parameters~\cite{guo2020parameter,zaken2021bitfit}; (ii) adjusting newly added parameters to the model or its input~\cite{li2021prefix,zhou2022learning,xin2024mmap}; (iii) applying low-rank factorization techniques to the parameters that require updates~\cite{hu2021lora,karimi2021compacter,haoconsolidator,liu2024m2istmultimodalinteractivesidetuning,xin2024vmt}. Some pioneering works like ETRIS~\cite{xu2023bridging} and DARA~\cite{liu2024dara} sought to utilize adapters to adapt pre-trained models to referring image segmentation and referring expression comprehension, respectively.
However, their proposed modules like Bridger~\cite{xu2023bridging} and RA~\cite{liu2024dara} are insufficient for capturing the complexity of multi-scale local visual features.

\begin{figure*}[t]
\centering
\includegraphics[width=\textwidth]{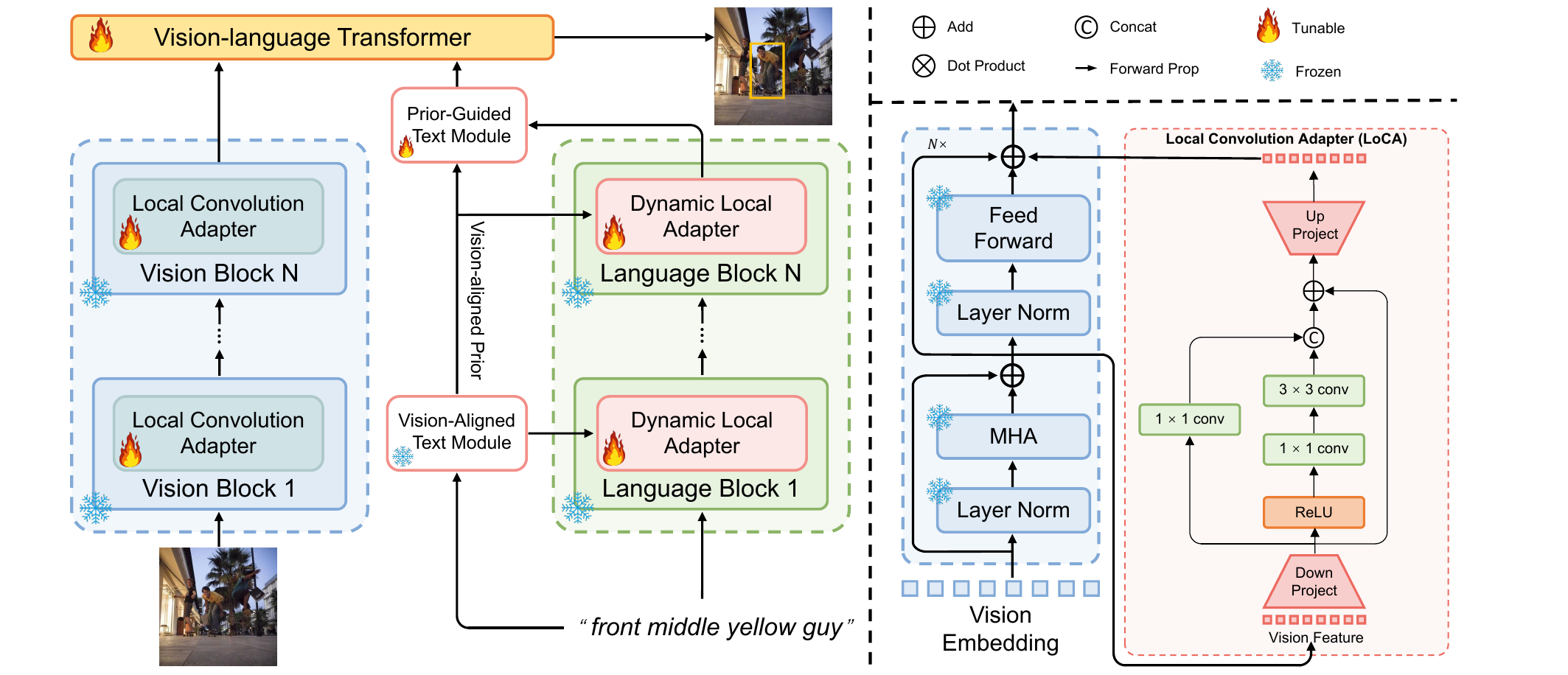}
\vspace{-6mm}
\caption{\textbf{Overall architecture of MaPPER.} MaPPER freezes the pre-trained vision encoder and language encoder. For the language branch, Dynamic Prior Adapters (DyPA) utilize aligned priors generated from the Vision-aligned Prior Module to enable efficient modal alignment and adaptation. For the language branch, Local Convolution Adapters (LoCA) integrate local visual features the global prior (pre-trained visual knowledge) from the visual encoder. 
Moreover, the Prior-guided Text module for promoting the multimodal alignment.}
\vspace{-4mm}
\label{fig:overview}
\end{figure*}

\section{Methodology}

\subsection{Framework Overview}
\label{Framework Overview}

The overall framework of the proposed MaPPER is illustrated in Figure \ref{fig:overview}. Our approach freezes the pre-trained backbone, ensuring parameter efficiency. 
This framework consists of two distinct efficient tuning modules. The first module, known as the Dynamic Prior Adapter, utilizes aligned prior generated from the Vision-aligned Prior Module to enable efficient modal alignment and adaptation. The second module, referred to as the Local Convolution Adapter module, integrates local visual features into global prior (pre-trained visual knowledge) from the visual encoder, thereby regularizing the whole visual perception. Finally, the complete textual features, alongside aligned prior, are inputted into the 
Prior-guided Text module for promoting the multimodal alignment.

\subsection{Text \& Image Feature Extraction}
\textbf{Text Encoder.} The REC task relies heavily on word-level understanding due to its concise linguistic expression format, such as "front middle yellow guy", to convey referring information. Owing to its bi-directional encoder representations and the masked language modeling, BERT~\cite{devlin2018bert} excels in word-level understanding, making it suitable for text encoding in REC domain. Given the input referring expression $T$, the text expression is firstly converted into a one-hot vector. Subsequently, each one-hot vector is tokenized into a series of linguistic tokens. A special \texttt{[CLS]} token is prefixed to the sequence, and the sequence of tokens is then fed into a stack of 12 transformer encoder layers to progressively capture and model the intricate language tokens.
\label{text encoder:bert}

\noindent
\textbf{Visual Encoder.} Our work adopts the transformer-based DINOv2-B/14~\cite{oquab2023dinov2} as the visual backbone. The model involves training the Vision Transformer (ViT) model~\cite{dosovitskiy2020vit} on the extensive LVD-142M dataset, utilizing a self-supervised learning strategy. This approach equips the model with the ability to extract powerful visual features, which in turn delivers impressive performance across various downstream tasks. Given an input image $\bm{I}_0 \in \mathbbm{R}^{H_0\times W_0\times3}$, the image is initially divided into $N$ non-overlapping patches, which are then linearly projected into $D$-dim patch embeddings $\bm{I}_p \in \mathbbm{R}^{N\times D}$. Meanwhile, a learnable \texttt{[CLS]} token is prepended to $\bm{I}_p$, 
producing $\bm{I} \in \mathbbm{R}^{(N+1)\times D}$.

Considering the substantial number of parameters, we opt to freeze visual and text encoders during the fine-tuning process. This strategy allows for a more efficient allocation of computational resources and focuses the learning on the adjustments of other modules.

\subsection{Prior-guided Text Understanding}
\label{Vision-Conditioned Text Understanding}
As detailed in section~\ref{text encoder:bert}, the pre-training mechanism of BERT makes it ideal for the REC task, which has a relatively high word-level understanding. However, BERT lacks alignment with vision in the pre-training process, and we introduce a Vision-aligned Prior Module to generate a vision-aligned prior. The prior serves for better adjusting BERT encoder, and promoting the interaction of text and vision features.

\noindent
\textbf{Vision-aligned Prior Module (VAP).} The core of VAP to a produce vision-aligned prior for the REC domain.
Considering that CLIP~\cite{radford2021learning} model inherently has the ability to align visual with text feature, we used the frozen CLIP followed by a mapping layer $M$ as the VAP module. Given the text input $\boldsymbol{t}$, the vision-aligned prior $\boldsymbol{p}$ can be formulated as follows:
\begin{equation}
\boldsymbol{p} = M(\operatorname{CLIP_f}(\boldsymbol{t})).
\end{equation}
where the $\operatorname{CLIP_f}$ denotes the frozen CLIP backbone.

\begin{figure}[t]
\centering
\includegraphics[width=1.0\columnwidth]{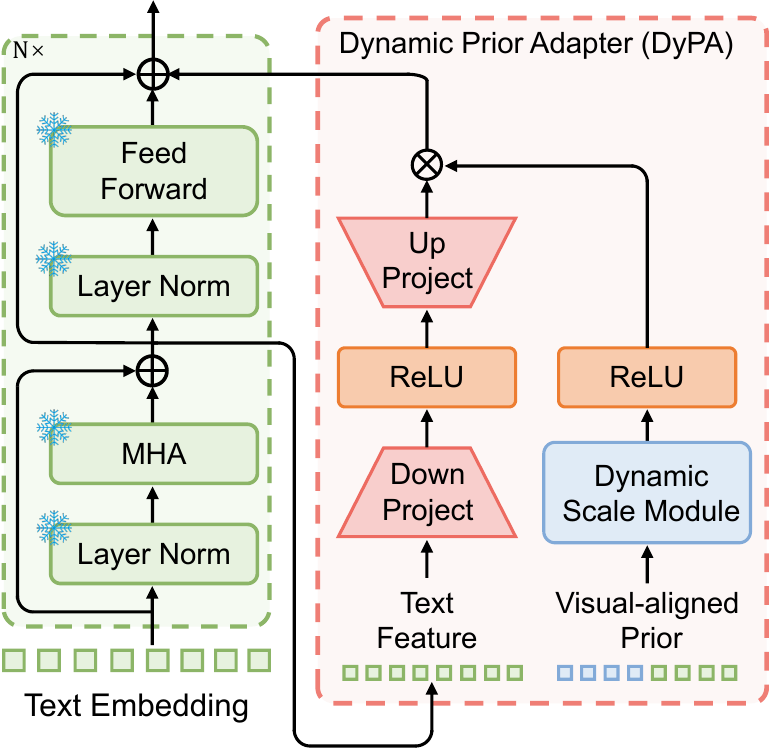}
\caption{The structure of the Dynamic Prior Adapter.
}
\label{fig:dypa}
\end{figure}

\noindent 
\textbf{Dynamic Prior Adapter (DyPA).} To dynamically bridge the gap between the pre-trained BERT model and the complex REC task, we introduce the Dynamic Prior Adapter, which operates in parallel with the text encoder, as shown in Figure \ref{fig:dypa}. DyPA comprising four module: a dynamic scale module (DS), a downward projection with parameters $\boldsymbol{W}_{down}^t \in \mathbbm{R}^{r \times d}$, a ReLU activation layer, and an upward projection with parameters $\boldsymbol{W}_{up}^t \in \mathbbm{R}^{d \times r}$.

Specifically, we adopt the DS module for integrating the vision-aligned prior $\boldsymbol{p}$ to different layers in the BERT encoder. The module generates scale factors $S_f$ using a scoring weight matrix $\boldsymbol{W}_{s} \in \mathbbm{R}^{1 \times d}$, eliminating manual hyper-parameter tuning. Given the prior $\boldsymbol{p}$, the dynamic scaling factor can be formulated as follows:
\begin{equation}
S_f = \operatorname{ReLU}\left ( \boldsymbol{p}\boldsymbol{W}_{s} \right ).
\end{equation}

\noindent
The downward projection and the upward projection are connected by a ReLU function. In one text encoder layer, the downward projection layer receives processed language tokens $\boldsymbol{x_t}$ from the Multi-head Attention (MHA) layer as input and produces adapted. 
In general, the output of DyPA $\boldsymbol{x_t}^{\prime}$ can be described as 
\begin{equation}
\boldsymbol{x_t}^{\prime}=S_f \times \left [ \left ( \operatorname{ReLU} \left ( \boldsymbol{x_t}\boldsymbol{W}_{down}^t \right )\right ) \boldsymbol{W}_{up}^t\right ].
\end{equation}

\noindent
DyPA utilizes the vision-aligned prior $p$ to dynamically regularize the feed-forward during adapter tuning. To mitigate the influence of Adapter outputs during the initial stages of model training, we initialize $\boldsymbol{W}_{up}^t$ to zero.

\noindent 
\textbf{Prior-guided Text Module (PGT).} Through the design of the DyPA module, we efficiently fine-tune the BERT model to produce fine-grained aligned text features for the REC tasks. In order to promote the interaction of text and vision features for the Multimodal Interactive Module in Sec.\ref{Multimodal Interactive Module}, we propose a Prior-Guided Text Module, fusing the prior $\boldsymbol{p} \in \mathbbm{R}^{N_t \times C_p}$ into the text features $\boldsymbol{t} \in \mathbbm{R}^{N_t \times C_t}$ generated by the BERT encoder. To achieve this, we employ a projection layer, denoted by $Proj \in \mathbbm{R}^{C_p \times C_t}$, to map the prior $\boldsymbol{p}$ onto a transformed representation $\boldsymbol{p'}$. This projection is specifically designed to align the dimensions of the prior with the text features. In order to streamline the process, we concatenate $\boldsymbol{t}$ with the transformed priors $\boldsymbol{p'}$ to get the final text feature $\boldsymbol{f_t}$ integrated with the vision-aligned prior.

\begin{equation}
\begin{aligned}
\boldsymbol{p'} = Proj(\boldsymbol{p'}),\\
\boldsymbol{f_t} = \text{Concat} [\boldsymbol{p'},\boldsymbol{t}].
\end{aligned}
\end{equation}

\subsection{Global \& Local Visual Perception}
\label{Global Local Visual Perception}
For visual perception in the REC task, local features and global representations are important counterparts. Although pre-trained DINOv2 can provide powerful and robust visual features to achieve promising performance, the task-specific visual attention in the REC task often focuses on localized areas of uncertain size in images, which have been visualized in Figure \ref{fig:visualizations}. 

\noindent
\textbf{Local Convolution Adapter (LoCA).} To further facilitate the visual perception ability of DINOv2 for the REC task, we propose a Local Convolution Adapter (LoCA) module to adjust the visual foundation models. LoCA introduces the multi-scale local information to further enhance visual perception. The local convolution adapter consists of a down-projection layer $\boldsymbol{W}_{down}^v$, a multi-scale convolution module, a ReLU activation layer, and the up-projection layer $\boldsymbol{W}_{up}^v$. 

Specifically, in one visual encoder layer, the downward projection layer receives processed visual tokens $\boldsymbol{x_v}$ from the Multi-head Attention (MHA) layer as input and produces adapted. The multi-scale convolution module consists of two parallel convolutional paths of multi-scale (1×1, 3×3). The 1×1 convolution is strategically placed before the 3×3 convolutions to reduce channel dimension. This design and the bottleneck structure make the local convolution adapter still lightweight. The outputs of the multi-scale convolutional paths are concatenated to form the local feature $f_{loc}$. 

\begin{equation}
\begin{aligned}
&f_v = \operatorname{ReLU}\boldsymbol(x_v\mathbf{W}_{down}^v), \\
&f_{v1} = \text{Conv}_{1 \times 1}(f_v), \\
&f_{v2} = \text{Conv}_{3 \times 3}(\text{Conv}_{1 \times 1}(f_v)), \\
&f_{loc} = \text{Concat}[f_{v1},f_{v2}]. 
\end{aligned}
\end{equation}

\noindent
before the up-projection, a skip connection operates in parallel with the multi-scale convolution module.

\begin{equation}
\begin{aligned}
&f_{loc'} = f_{loc}+f_v, \\
&f_{loc} = \boldsymbol(f_{loc'}\mathbf{W}_{up}^v). 
\end{aligned}
\end{equation}

\noindent
\textbf{Global and Local Visual Integration.} 
To augment the DINOv2 backbone with multi-scale local visual perception on the REC task, we integrate the Local Convolution Adapter (LoCA) in parallel with the MLP layer within the transformer block. By the concise design, LoCA module adds multi-scale local prior into the DINOv2 model for the REC task. The output of each adapted transformer block can be described as:

\begin{equation}
\begin{aligned}
&v_l^{mha} = \operatorname{MHA}(\operatorname{LN}(v_{l-1}))+v_{l-1}, \\
&v_l = \operatorname{MLP}(\operatorname{LN}(v_l^{mha}))+s\cdot f_{loc} + v_l^{mha}. 
\end{aligned}
\end{equation}

\noindent
where $s$ is the scaling factor, and the $v_{l-1}$ represents the previous layer output.

\begin{table*}[!t]
\centering

\small
\setlength{\tabcolsep}{2.2pt}

\resizebox{1\textwidth}{!}{
\begin{tabular}{lllccccccccccll}
\toprule
    
\multicolumn{1}{c|}{\multirow{2}{*}{Methods}}&\multicolumn{1}{c|}{\multirow{2}{*}{Venue}} & \multicolumn{1}{c|}{Tuned/Total} & \multicolumn{3}{c|}{RefCOCO} & \multicolumn{3}{c|}{RefCOCO+} & \multicolumn{3}{c}{RefCOCOg} & & \\

  \multicolumn{1}{c|}{}& \multicolumn{1}{c|}{}& \multicolumn{1}{c|}{param.} & val & testA & \multicolumn{1}{c|}{testB} & val & testA & \multicolumn{1}{c|}{testB} & val-g & val-u & \multicolumn{1}{c}{test-u} & & \\ \midrule

\rowcolor{gray!20}\textbf{Full Fine-tuning}   & & &    &  &  &  && &  & &   & &\\
\midrule

\multicolumn{1}{c|}{MAttNet~\cite{yu2018mattnet}}    &\multicolumn{1}{c|}{CVPR'18} & \multicolumn{1}{c|}{100\%} & 76.65 & 81.14 &  \multicolumn{1}{c|}{69.99} & 65.33 & 71.62 & \multicolumn{1}{c|}{56.02} & - & 66.58 & 67.27  & &\\
\multicolumn{1}{c|}{RvG-Tree~\cite{hong2019learning}}   &\multicolumn{1}{c|}{TPAMI'19} & \multicolumn{1}{c|}{100\%} & 75.06 & 78.61 &  \multicolumn{1}{c|}{69.85} & 63.51 & 67.45 & \multicolumn{1}{c|}{56.66} & - & 66.95 & 66.51  & &\\

\multicolumn{1}{c|}{NMTree~\cite{liu2019nmtree}}   &\multicolumn{1}{c|}{ICCV'19} & \multicolumn{1}{c|}{100\%} & 76.41 & 81.21 &  \multicolumn{1}{c|}{70.09} & 66.46 & 72.02 & \multicolumn{1}{c|}{57.52} & 64.62 & 65.87 & 66.44  & &\\ 


\multicolumn{1}{c|}{FAOA~\cite{yang2019fast}}  &\multicolumn{1}{c|}{ICCV'19} & \multicolumn{1}{c|}{100\%} & 72.54 & 74.35 &  \multicolumn{1}{c|}{68.50} & 56.81 & 60.23 & \multicolumn{1}{c|}{49.60} & 56.12 & 61.33 & 60.26  & &\\
\multicolumn{1}{c|}{ReSC-Large~\cite{yang2020improving}}   &\multicolumn{1}{c|}{ECCV'20} & \multicolumn{1}{c|}{100\%} & 77.63 & 80.45 & \multicolumn{1}{c|}{72.30} & 63.59 & 68.36 & \multicolumn{1}{c|}{56.81} & 63.12 & 67.30 & 67.20  & &\\ 
\multicolumn{1}{c|}{TransVG~\cite{deng2021transvg}}   & \multicolumn{1}{c|}{ICCV'21} & \multicolumn{1}{c|}{100\%}&  80.32 & 82.67 & \multicolumn{1}{c|}{78.12} & 63.50 & 68.15 & \multicolumn{1}{c|}{55.63} & 66.56 & 67.66 & 67.44  & &\\
\multicolumn{1}{c|}{QRNet~\cite{ye2022shifting}}     & \multicolumn{1}{c|}{CVPR'22} & \multicolumn{1}{c|}{100\%}   & 84.01  & 85.85  & \multicolumn{1}{c|}{\textbf{82.34}}  & 72.94  & 76.17  & \multicolumn{1}{c|}{63.81}  & 71.89  & 73.03  &72.52  & &\\ 

\multicolumn{1}{c|}{Dynamic-MDETR $^\dagger$~\cite{shi2023dynamic}}  & \multicolumn{1}{c|}{TPAMI’23}& \multicolumn{1}{c|}{100\%}  & \underline{85.97}  & \underline{88.82}  & \multicolumn{1}{c|}{80.12}  & \underline{74.83}  & \textbf{81.70}  & \multicolumn{1}{c|}{63.44}  & 72.21& 74.14  & 74.49   & &\\
\multicolumn{1}{c|}{PFOS~\cite{sun2022proposal}}   &\multicolumn{1}{c|}{TMM’22} & \multicolumn{1}{c|}{100\%} & 77.37 & 80.43 & \multicolumn{1}{c|}{72.87} & 63.74 & 68.54 & \multicolumn{1}{c|}{55.84} & 61.46 & 67.08 & 66.35  & &\\
\multicolumn{1}{c|}{SeqTR~\cite{zhu2022seqtr}}   & \multicolumn{1}{c|}{ECCV'22} & \multicolumn{1}{c|}{100\%}   & 81.23  & 85.00  & \multicolumn{1}{c|}{76.08}  & 68.82  & 75.37  & \multicolumn{1}{c|}{58.78} & - & 71.35  & 71.58    & &\\  
\multicolumn{1}{c|}{Word2Pix~\cite{zhao2022word2pix}}     & \multicolumn{1}{c|}{TNNLS’22} & \multicolumn{1}{c|}{100\%}   & 81.20  & 84.39  & \multicolumn{1}{c|}{78.12}  & 69.46  & 76.81  & \multicolumn{1}{c|}{61.57} & -& 70.81  & 71.34      & &\\
\multicolumn{1}{c|}{YORO$^\dagger$~\cite{ho2023yoro}}   & \multicolumn{1}{c|}{ECCV'22} & \multicolumn{1}{c|}{100\%}  & 82.90  & 85.60  & \multicolumn{1}{c|}{77.40}  & 73.50  & 78.60  & \multicolumn{1}{c|}{64.90} & - & 73.40  & 74.30    & &\\

\multicolumn{1}{c|}{CLIP-VG~\cite{xiao2023clip}}  & \multicolumn{1}{c|}{TMM'23} & \multicolumn{1}{c|}{100\%}   & 84.29  & 87.76  & \multicolumn{1}{c|}{78.43}  & 69.55  & 77.33  & \multicolumn{1}{c|}{57.62} & 72.64 & 73.18  & 72.54  & &\\  
\multicolumn{1}{c|}{JMRI~\cite{zhu2023jmri}}  & \multicolumn{1}{c|}{TIM'23} & \multicolumn{1}{c|}{100\%}    & 82.97 & 87.30  & \multicolumn{1}{c|}{74.62}  & 71.17  & 79.82  & \multicolumn{1}{c|}{57.01} & 69.32  & 71.96  & 72.04   & &\\

\multicolumn{1}{c|}{MGCross~\cite{miao2023self}}      & \multicolumn{1}{c|}{TIP'24} & \multicolumn{1}{c|}{100\%}   & 85.10  & 88.23  & \multicolumn{1}{c|}{80.08}  & 74.44  & 79.48  & \multicolumn{1}{c|}{\underline{65.21}} & \underline{74.50} & \textbf{77.25}   & \underline{75.78}     & &\\

\multicolumn{1}{c|}{TransCP~\cite{tang2023context}}      & \multicolumn{1}{c|}{TPAMI'24} & \multicolumn{1}{c|}{100\%}   & 84.25  & 87.38  & \multicolumn{1}{c|}{79.78}  & 73.07  & 78.05  & \multicolumn{1}{c|}{63.35} & 72.60 & -   & -     & &\\

\multicolumn{1}{c|}{ScanFormer~\cite{su2024scanformer}}      & \multicolumn{1}{c|}{CVPR'24} & \multicolumn{1}{c|}{100\%}   & 83.40  & 85.86  & \multicolumn{1}{c|}{78.81}  & 72.96  & 77.57  & \multicolumn{1}{c|}{62.50} & 74.10 & -  & 74.14     & &\\

\midrule
\rowcolor{gray!20}\textbf{Parameter-efficient Transfer Learning}   & &   &  &  &  &  && &  & &   & &\\
\midrule

\multicolumn{1}{c|}{DARA~\cite{liu2024dara}}  & \multicolumn{1}{c|}{Arxiv'24} & \multicolumn{1}{c|}{1.63\%}   & 81.16   & 82.76  & \multicolumn{1}{c|}{76.72}  & 65.58  & 69.83  & \multicolumn{1}{c|}{57.22} & 67.21 & 69.22 & 67.67    & &\\

\multicolumn{1}{c|}{\textbf{MaPPER (Ours)}}   & \multicolumn{1}{c|}{-} & \multicolumn{1}{c|}{\textbf{1.41\%}} & \textbf{86.03} & \textbf{88.90} & \multicolumn{1}{c|}{\underline{81.19}} &  \textbf{74.92}& \underline{81.12} & \multicolumn{1}{c|}{\textbf{65.68}}& \textbf{74.60} & \underline{76.32} & \textbf{75.81}  & &\\
\bottomrule

\end{tabular}
}

\caption{\textbf{Comparison with latest SOTA methods on RefCOCO/+/g for visual grounding. $\dagger$ indicates that all of the RefCOCO/+/g training data has been used during pre-training.} "Tuned/Total param." is the average percentage of tuned parameters in backbone. We highlight the \textbf{best} and the \underline{second-best} results.}
\vspace{-3mm}
\label{Table:comparisons with SOTA}
\end{table*}

\subsection{Multimodal Interactive Module}
\label{Multimodal Interactive Module}
We have implemented a transformer~\cite{vaswani2017attention} architecture that seamlessly integrates multimodal embeddings to forecast the bounding box of the referenced object. Specifically, the adapted vision embeddings $\bm{f}_v \in \mathbbm{R}^{N_v \times C_v}$ and language embeddings $\bm{f}_l \in \mathbbm{R}^{N_l \times C_l}$ are first projected into a common space of joint embeddings $\bm{f'}_v \in \mathbbm{R}^{N_v \times C_p}$ and $\bm{f'}_l \in \mathbbm{R}^{N_l \times C_p}$, both with a unified channel size. Followed by TransVG~\cite{deng2021transvg} and DARA~\cite{liu2024dara}, these joint embeddings, along with a learnable \texttt{[REG]} token, are processed through a series of six transformer encoder layers, to fuse the cross-modality embeddings. Finally, a prediction head, implemented as a Multi-layer Perceptron with two 256-dimensional hidden layers and a linear output layer, takes the \texttt{[REG]} token as input and projects it onto the 4-dimensional coordinates for defining the bounding box.

we \section{Experiments}
\label{sec:experiments}

\begin{table*}[!t]
\centering

\small
\setlength{\tabcolsep}{4.5pt}

\begin{tabular}{l|c|ccc|ccc|ccc}
\toprule
    
\multirow{2}{*}{Methods} & \multicolumn{1}{c}{Params.$\downarrow$}  & \multicolumn{3}{c|}{RefCOCO} & \multicolumn{3}{c|}{RefCOCO+} & \multicolumn{3}{c}{RefCOCOg} \\

 & (M) & val & testA & testB & val & testA & testB & val-g & val-u & test-u \\ \midrule
 \textcolor{gray}{Fully fine-tuning} & \textcolor{gray}{196} & \textcolor{gray}{85.31} & \textcolor{gray}{87.80} & \textcolor{gray}{81.03} & \textcolor{gray}{74.57} & \textcolor{gray}{80.22} & \textcolor{gray}{65.31} & \textcolor{gray}{73.76} & \textcolor{gray}{74.22} & \textcolor{gray}{75.02}  \\ \midrule

Adapter \cite{houlsby2019adapter} & 4.09  & 84.23 & 86.76 & 79.98 & 73.76 & 79.91 & 65.14 & 72.37 & 74.19 & 74.25  \\

LoRA \cite{hu2021lora} & 3.25   & 83.13 & 85.51 & 78.32 & 73.66 & 78.73 & 64.85 & 73.67 & 74.66 & 74.83 \\

AdaptFormer \cite{chen2022adaptformer} & 4.09 & 84.75 & 86.14 & 79.73 & 73.05 & 79.63 & 65.26 & 72.19 & 73.93 & 74.36 \\

CM Adapter \cite{jiang2022cmadapter} & 4.09 & 85.84 & 86.49 & 79.67 & 74.06 & 79.91 & 64.27 & 73.61 & 73.54 & 74.19 \\

MRS-Adapter \cite{yuan2023mrsadapter} & 2.98  & 83.92 & 85.06 & 78.52 & 71.13 & 78.38 & 63.13 & 72.42 & 73.26 & 72.92  \\

\midrule

\rowcolor{gray!20}
\textbf{MaPPER} & \textbf{2.77} & \textbf{86.03} & \textbf{88.90} & \multicolumn{1}{c|}{\textbf{81.19}} &  \textbf{74.92}& \textbf{81.12} & \multicolumn{1}{c|}{\textbf{65.68}}& \textbf{74.60} & \textbf{76.32} & \textbf{75.81}  \\
\bottomrule

\end{tabular}
\caption{\textbf{Comparison with PETL methods using the DINO-B Backbone on RefCOCO, RefCOCO+ and RefCOCOg.} "Param." indicates the number of tunable parameters in the pre-trained encoders. To ensure fairness, we kept the original parameter settings from previous methods. }
\vspace{-3mm}
\label{Table:comparisons with PETL}
\end{table*}

\subsection{Experimental Setup}
\noindent \textbf{Datasets and Evaluation Metrics.} We validate our method on three widely-used REC benchmarks: RefCOCO~\cite{yu2016refcoco}, RefCOCO+~\cite{yu2016refcoco}, and RefCOCOg~\cite{mao2016refcocogg,nagaraja2016refcocogu}. We follow the previous research that employs top-1 accuracy (\%) as the evaluation metric. Specifically, a prediction is deemed accurate only when its IoU exceeds or equals 0.5. In addition to Precision@0.5, we also report the number of tunable parameters in the pre-trained encoders  to compare the fine-tuning efficiency with traditional full fine-tuning and other PETL methods.

\noindent \textbf{Implementation Details.} The vision encoder is initialized with DINOv2-B/14~\cite{oquab2023dinov2}, while the language encoder uses BERT-base~\cite{devlin2018bert}. The resolution of the input image is 518×518. Both the DINOv2-B/14 model and the BERT-base model process tokens with a feature dimension of 768. The  Multimodal Interactive Module uses Xavier initialization. DyPA are initialized with Kaiming normal initialization and inserted into the transformer layers for the language encoder. The bottleneck dimension $C_d$ for DyPA is 32. For LoCA, the 1×1 convolution before the 3×3 convolution reduces the channel to 24. The output dimensions of the two convolutional paths are 192 and 96, so the input dimension of the these convolutional paths is 288. For fair comparisons, PETL methods in Table \ref{Table:comparisons with PETL} use the same base architecture, and keeping the vision and language encoder fixed.

\subsection{Main Results}
We conducted a comprehensive comparison between our proposed MaPPER model and a series of previous referring expression comprehension (REC) methods. The main experimental results are presented in Table~\ref{Table:comparisons with SOTA}, from which we can observe that: MaPPER achieves the best accuracy while ensuring parameter efficiency among all methods, thus validating its effectiveness and efficiency.

\noindent
\textbf{Effectiveness.} As Table~\ref{Table:comparisons with SOTA} shown, on the three commonly challenging benchmarks, MaPPER outperforms all traditional full fine-tuning methods. Compared to DARA~\cite{liu2024dara}, a parameter-efficient transfer learning method, we achieves best results on the three benchmarks. Notably, even compared to some methods that are pre-trained on the the RefCOCO/+/g (indicated by $\dagger$ in Table \ref{Table:comparisons with SOTA}), our MaPPER model achieves the highest scores across all evaluation tasks, with particularly strong performance on the RefCOCO+, which present greater challenges compared to RefCOCO.

\noindent
\textbf{Efficiency.} Table~\ref{Table:comparisons with SOTA} clearly illustrates that MaPPER not only achieves the best performance, but also highlights its huge advantages in parameter efficiency. MaPPER reduced the tunable backbone parameters by 98.59\% compared to the traditional full fine tuning method. Compared to the PETL method DARA~\cite{liu2024dara}, MaPPER has also lower tunable parameters.

\begin{table}[!t]
\centering

\small
\setlength{\tabcolsep}{2.1pt}

\begin{tabular}{l|c|c|ccc}
\toprule
    
 \multirow{2}{*}{\#} & \multirow{2}{*}{Local Conv. Adapter} & \multicolumn{1}{c|}{Params.} & \multicolumn{3}{c}{RefCOCO} \\

 & & (M) & val & testA & testB \\ \midrule

(a)  &  & 0 &  82.37 & 84.13 & 77.57 \\ 
\midrule
(b) & \checkmark &  1.58 & 84.28 & 86.02 & 79.38 \\

\bottomrule

\end{tabular}
\vspace{-2mm}
\caption{\textbf{Effectiveness of Local Convolution Adapter (LoCA)} for the visual branch. Note the ablation study without adding any component in the text branch, and we freeze the text encoder. (a) represents freezing both the text and visual branche. }
\vspace{-4mm}
\label{Table:ablation on visual}
\end{table}
\subsection{Comparison with Other PETL Methods}

We conduct experiments comparing our MaPPER with other parameter-efficient tuning methods using DINOv2-Base as the backbone. To ensure fairness, we retain the original parameter settings from previous methods and adjust the bottleneck to achieve comparable parameter counts. Table~\ref{Table:comparisons with PETL} illustrates that MaPPER outperforms other PETL methods on all three benchmarks, and even performs better than fully fine-tuning. This highlights the effectiveness of MaPPER in adapting pre-trained knowledge for the REC domain. Through introducing vision-aligned prior, MaPPER enhance the modeling of the vision-text alignment capability. Furthermore, inserting Local Convolution Adapters into DINOv2, making it more suitable for REC tasks with enhanced local visual perception. Previous PETL methods lack these functionalities, rendering them less effective for REC tasks. To summarize, by the specific design for the REC domain, MaPPER achieves superior performance with only \textbf{2.77} million parameters.

\subsection{Ablation Study}

\textbf{Effectiveness of Local Convolution Adapter.} We assess the impact of the Local Convolution Adapter (LoCA) by performing an ablation study and reporting the results on RefCOCO validation and test datasets. From Table \ref{Table:ablation on visual}, it is evident that introducing the LoCA yields a great improvement, increasing the average performance to 1.87\%. This indicates that the LoCA enhances the visual perception of DINOv2 with local visual feature.

\begin{figure*}[t]
\centering
\includegraphics[width=\textwidth]{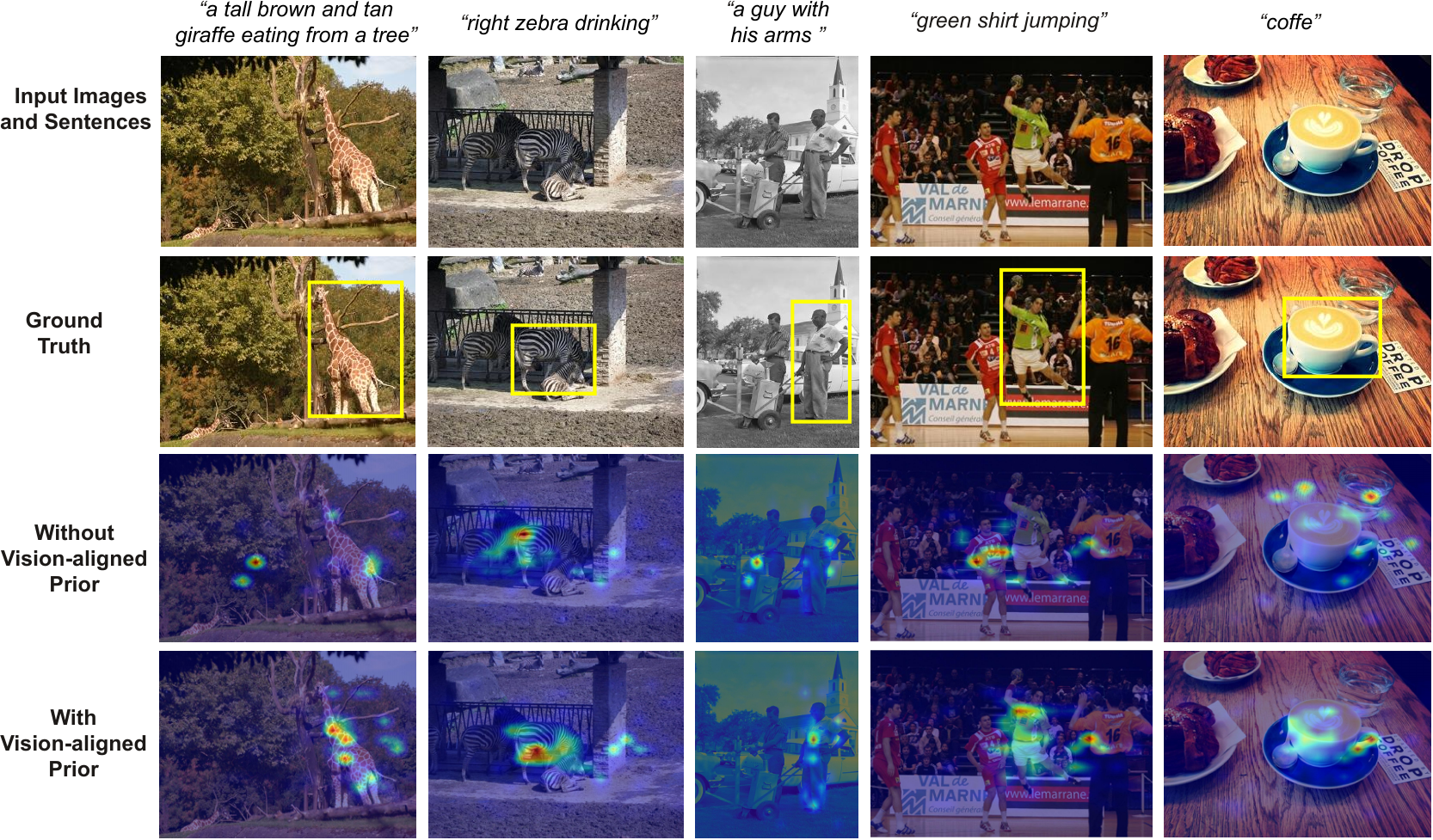}
\vspace{-6mm}
\caption{Visualizations of attention maps from the Multimodal Interactive Module for validating the effect of the vision-aligned prior.}
\vspace{-4mm}
\label{fig:visualizations}
\end{figure*}

\begin{table}[!t]
\centering

\small
\setlength{\tabcolsep}{2.1pt}

\begin{tabular}{l|c|c|ccc}
\toprule
    
 \multirow{2}{*}{\#} & \multirow{2}{*}{Multi-scale size} & \multicolumn{1}{c|}{Params.} & \multicolumn{3}{c}{RefCOCO} \\

 & & (M) & val & testA & testB \\ \midrule

(a)  & 1×1 & 1.48 &  83.51 & 85.35 & 78.56 \\
\rowcolor{gray!20}
(b) & 1×1 3×3&  1.58 & 84.28 & 86.02 & 79.38 \\
(c) & 1×1 3×3 5×5&  1.70 & 83.98 & 85.72 & 79.02 \\

\bottomrule

\end{tabular}
\vspace{-2mm}
\caption{\textbf{Effectiveness of multi-scale size} for the visual branch.}
\vspace{-4mm}
\label{Table:multi-scale}
\end{table}

\begin{table}[!t]
\centering

\small
\setlength{\tabcolsep}{2.1pt}

\resizebox{0.48\textwidth}{!}{
\begin{tabular}{l|ccc|ccccc}
\toprule
    
 \multirow{2}{*}{\#} & \multicolumn{1}{c}{Adapter} & \multicolumn{1}{c}{Adapter} & \multicolumn{1}{c|}{Params.} & \multicolumn{3}{c}{RefCOCO} \\

 & w/o $\boldsymbol{p}$ & w. $\boldsymbol{p}$ (DyPA)  & (M) & val & testA & testB \\ \midrule

(a)  &  &   & 1.58 & 84.28 & 86.02 &79.38\\ 
\midrule
(b) & \checkmark &  & 1.73(+0.15) & 84.78 & 86.62 & 79.89 \\
(c) &  & \checkmark  & 1.79(+0.21)  & 85.32 & 87.62 & 80.12 \\
\midrule
\midrule
\multirow{2}{*}{\#} & \multicolumn{1}{c}{Adapter} & \multirow{2}{*}{PGT} & \multicolumn{1}{c|}{Params.} & \multicolumn{3}{c}{RefCOCO} \\

 & w. $\boldsymbol{p}$ (DyPA)&  & (M) & val & testA & testB \\ \midrule

(d)&   & \checkmark & 2.56(+0.98) & 85.76 & 87.88 & 80.98 \\
\rowcolor{gray!20}
(f)  & \checkmark & \checkmark & 2.77(+1.19)  & \textbf{86.03} & \textbf{88.90} & \textbf{81.19} 
\\
\bottomrule

\end{tabular}
}
\caption{\textbf{Effectiveness of the Vision-Prior
for the text branch}. Note (a) represents freezing the text encoder while tuning the LoCA in the visual encoder, and the \textbf{LoCA included in Params.}. (b) represents using the vanilla adapter without $\boldsymbol{p}$. (d) represents only using the PGT without any adapters.}
\vspace{-4mm}
\label{Table:ablation on text}
\end{table}

\noindent
\textbf{Effect of Multi-scale Size for Visual Branch.}
To further verify the effect of local visual information, we perform the attempts 
of using only a single-size convolution kernel (1×1), and three scales (1×1 3×3 5×5).
Table \ref{Table:multi-scale} indicates that it is difficult for an adapter with a single-size convolution kernel (a) to perform well for the REC. Local Features are too fine-grained (c) are also not optimal. In contrast, appropriate multi-scale (b)
provide proper local information, thus achieving the best performance.

\noindent
\textbf{Effect of the Vision-aligned Prior for Text Branch.}
From Table~\ref{Table:ablation on text}, we can see that: \textbf{(1)} Freezing the text encoder while only tuning local convolution adapter can also brings great performance (Table~\ref{Table:ablation on text} (a)); \textbf{(2)} it is crucial to obtain a dynamic scale with the vision-prior, the Dynamic Prior Adapter (DyPA) brings better performance compared to vanilla adapter fixing the scale to 1.0 (Table~\ref{Table:ablation on text} (b,c)); \textbf{(3)} by the design of Prior-guided Text Module (PGT), we further promote the interaction of text and vision features (Table~\ref{Table:ablation on text} (f)); \textbf{(4)} Incorporating the DyPA and PGT results in an average improvement of 1.02\% compared to only using DyPA. 

\vspace{-2mm}
\subsection{Qualitative Results}
\label{sec:Qualitative}
To investigate the impact of vision-aligned prior, we visualize the attention maps from the Multimodal Interactive Module under two strategies: with and without the vision-aligned prior. In the absence of the prior represents the text adapter without dynamic scale, and the prior-guided text module is not introduced. As shown in Fig. \ref{fig:visualizations}, referring expressions contain object appearance attributions, human actions, and spatial relationships. It is observable that the model can focus well on the local target region of the whole image with the vision-aligned prior. This indicates that vision-aligned prior enhancing the alignment ability of MaPPER.

\section{Conclusion}
\label{con}
In this study, we present an innovative Parameter-Efficient Transfer Learning (PETL) approach designed for multi-modal language grounding tasks, especially in referring expression comprehension. MaPPER enhances the adapters with multi-modal prior through the implementation of a simple yet effective fine-tuning strategy. We aims at improving both the effectiveness and efficiency of visual-text alignment, as well as enhancing visual perception by incorporating local visual semantics. The Dynamic Prior Adapter (DyPA) employs aligned prior to dynamically adjust the language encoder, while the Local Convolution Adapter (LoCA) introduces local visual features for enhancing the visual encoder. MaPPER not only surpasses the performance of fully fine-tuned models but also does more efficiently.

\section{Limitation}
\label{limit}
While our proposed method has shown enhanced efficiency, scalability, and parameter optimization in the realm of REC tasks, surpassing conventional fully fine-tuned models, our empirical inquiries have been confined to this specific domain. It is imperative for future research to broaden the validation scope encompass include variety range of other multi-modal tasks. Moreover, while our approach can effectively decrease the quantity of parameters necessitating training, thus conserving computational and storage resources, it still mandates a training process. As the frontier of multi-modal large-scale models progresses, there is a significant opportunity for future exploration into open-vocabulary zero-shot referring expression comprehension. This area of research could unveil innovative pathways and contribute to the evolution of models capable of comprehending and generating expressions without the constraint of prior training.

\bibliography{custom}

\appendix

\end{document}